\documentclass{article}

 \PassOptionsToPackage{numbers, compress}{natbib}

\usepackage[preprint]{nips_2018}

\usepackage{natbib}
\usepackage[utf8]{inputenc} 
\usepackage[T1]{fontenc}    
\usepackage{hyperref}       
\usepackage{url}            
\usepackage{subcaption}
\usepackage{booktabs}       
\usepackage{amsfonts}       
\usepackage{nicefrac}       
\usepackage{microtype}      

\usepackage[cmex10]{amsmath}
\usepackage{graphics}
\usepackage{graphicx}
\usepackage{verbatim}
\usepackage{epsfig,float}
\usepackage{cite}
\usepackage{algorithm}
\usepackage{algorithmic}
\usepackage{natbib}
\usepackage{times}
\usepackage{tikz}
\usetikzlibrary{arrows}
\usepackage{float}
\usepackage{amssymb}

\DeclareMathOperator*{\argmax}{arg\,max}

\title{Hierarchical Reinforcement Learning with Deep Nested Agents}

\author{
  Marc W.~Brittain\thanks{Website: marcbrittain.github.io (Marc Brittain)} \\
  Department of Aerospace Engineering\\
  Iowa State University\\
  Ames, IA 50011 \\
  \texttt{mwb@iastate.edu} \\
   \And Peng ~Wei\thanks{Website: http://www.aere.iastate.edu/~pwei/ (Peng Wei)}
 \\ Department of Aerospace Engineering\\
  Iowa State University \\
  Ames, IA 50011 \\
  \texttt{pwei@iastate.edu} \\
}

\begin{document}

\maketitle

\begin{abstract}
Deep hierarchical reinforcement learning has gained a lot of attention in recent years due to its ability to produce state-of-the-art results in challenging environments where non-hierarchical frameworks fail to learn useful policies. However, as problem domains become more complex, deep hierarchical reinforcement learning can become inefficient, leading to longer convergence times and poor performance. We introduce the Deep Nested Agent framework, which is a variant of deep hierarchical reinforcement learning where information from the main agent is propagated to the low level \textit{nested} agent by incorporating this information into the nested agent's state. We demonstrate the effectiveness and performance of the Deep Nested Agent framework by applying it to three scenarios in Minecraft with comparisons to a deep non-hierarchical single agent framework, as well as, a deep hierarchical framework.

\end{abstract}

\section{Introduction}

Deep reinforcement learning has recently attracted a large community due to recent successes like Deepmind's agent \textit{AlphaGO} \citep{alphago}. \textit{AlphaGO} was able to defeat the world champion Ke Jie in three matches of GO! in May 2017, which shows the capability that deep reinforcement learning has for solving challenging problems. A problem arises, however, in domains where the typical assumptions of reinforcement learning no longer hold. In environments where the Markovian assumption fails, actions taken early on can have long term effects that are unknown until the environment progresses for some time which lead to sub-optimal convergence in these domains. To alleviate this problem, a hierarchical formulation of reinforcement learning can be applied to create an Markovian environment consisting of a top-level and low-level agent (although more complex hierarchies consisting of many low-level agents are also possible).

The framework we propose draws inspiration from deep hierarchical reinforcement learning formulations, where there is a main agent as well as a low-level agent known as the \textit{nested agent}. The main agent is able to pass the information down to the nested agent in the form of state inclusion, where the information from the main agent can be a policy, task, objective, etc.

In this paper we utilize three non-markovian scenarios in Minecraft to demonstrate the efficiency and performance of the Deep Nested Agent framework, when compared to hierarchical reinforcement learning and non-hierarchical single agent reinforcement learning. Minecraft is a popular open-world generated environment that has recently attracted the attention of AI researchers due to the ability to create controlled scenarios. In this environment, we create our own controlled scenarios in what we call the \textit{arena} to test our Deep Nested Agent framework on. Figure 1 shows a screen-shot of the Minecraft arena that we created.  

The structure of this paper is as follows: in Section II, the background of reinforcement learning and deep Q-networks will be introduced. In Section III, a review of the current state-of-the-art hierarchical reinforcement learning approaches will be introduced. Section IV presents our proposed Deep Nested Agent framework to solve this problem. The numerical experiments as well as the results are shown in Section V, with Section VI concluding this paper.

\begin{figure}[H]
\centering
\includegraphics[width=9cm, height=7cm]{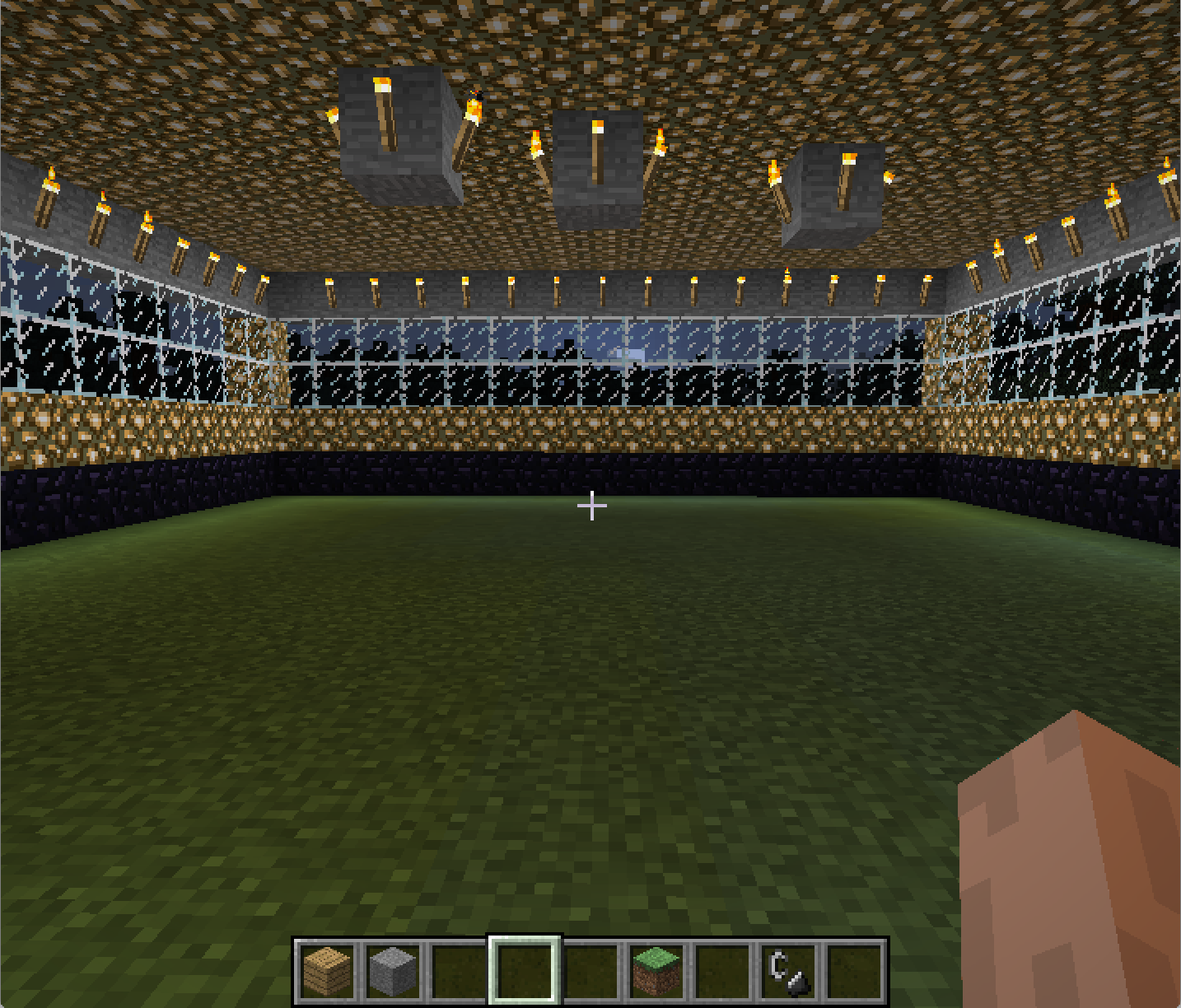} 
\caption{Arena created within the Minecraft environment.}
\end{figure}

\section{Background}
\subsection{Reinforcement Learning}
Reinforcement learning is one type of sequential decision making where the goal is to learn how to act optimally in a given environment with unknown dynamics. A reinforcement learning problem involves an environment, an agent, and different actions the agent can take in this environment. The agent is unique to the environment and we assume the agent is only interacting with one environment. Let $t$ represent the current time, then the components that make up a reinforcement learning problem are as follows:
\begin{itemize}
\item $S$ - The state space $S$ is a set of all possible states in the environment
\item $A$ - The action space $A$ is a set of all actions the agent can take in the environment
\item $r(s_{t},a_{t},s_{t+1})$ - The reward function determines how much reward the agent is able to acquire for a given $(s_{t},a_{t},s_{t+1})$ transition
\item $\gamma \in [0,1]$ - A discount factor determines how far in the future to look for rewards. As $\gamma \rightarrow 0$, only immediate rewards are considered, whereas, when $\gamma \rightarrow 1$, future rewards are prioritized.
\end{itemize}

$S$ contains all information about the environment and each element $s_{t}$ can be considered a \textit{snapshot} of the environment at time $t$. The agent accepts $s_{t}$ and with this, the agent then determines an action, $a_{t}$. By taking action $a_{t}$, the state is now updated to $s_{t+1}$ and there is an associated reward from making the transition from $s_{t} \rightarrow s_{t+1}$. How the state evolves from $s_{t} \rightarrow s_{t+1}$ given action $a_{t}$ is dependent upon the dynamics of the system, which is unknown. The reward function is user defined, but needs to be carefully designed to reflect the goal of the agent. Figure 1 shows the progression of a reinforcement learning problem.

\begin{figure}[h]
\begin{center}

\tikzstyle{int}=[draw, fill=blue!20, minimum size=2em]
\tikzstyle{init} = [pin edge={to-,thin,black}]

\begin{tikzpicture}[node distance=2.5cm,auto,>=latex']
    \node [int] (a) {Agent};
    \node (b) [left of=a,node distance=2cm, coordinate] {a};
    \node [int, pin={[init]above:$r$}] (c) [right of=a] {$s^{'}$};
    \node [coordinate] (end) [right of=c, node distance=2cm]{};
	\node [int] (d) [right of=c] {Agent};
    \path[->] (b) edge node {$s$} (a);
    \path[->] (a) edge node {$a$} (c);
    \path[->] (c) edge node {$s^{'}$} (d);
\end{tikzpicture}
\caption{Progression of a reinforcement learning problem within an environment.}
\end{center}
\end{figure}
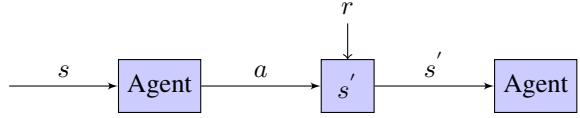

From this framework, the agent is able to learn the optimal decisions in each state of the environment by maximizing a cumulative reward function. We call the sequential actions the agent makes in the environment a \textit{policy}. Let $\pi$ represent some policy and $T$ represent the total time for a given environment, then the \textit{optimal policy} can be defined as:

\begin{equation}
\pi^{*}  = \argmax_{\pi} \mathrm{E}[\sum_{t = 0}^{T}(r(s_{t},a_{t},s_{t+1})|\; \pi) ].
\end{equation}

If we define the reward for actions we deem "optimal" very high, then by maximizing the total reward, we have found the optimal solution to the problem.

\subsection{Q-Learning}
One of the most fundamental reinforcement learning algorithms is known as Q-learning. This popular learning algorithm was introduced by \citet{watkins1989learning} and the goal is to maximize a cumulative reward by selecting an appropriate action in each state. The idea of Q-learning is to estimate a value $Q$ for each state and action pair $(s,a)$ in an environment that directly reflects the future reward associated with taking such an action from this state. By doing this, we can extract the policy that reflects the optimal actions for an agent to take. The policy can be thought of as a mapping or a look-up table, where at each state, the policy tells the agent which action is the best one to take. During each learning iteration, the Q-values are updated as follows:
\begin{equation}
Q(s_{t},a_{t}) \leftarrow Q(s_{t},a_{t}) + \alpha(r + \gamma\max_{a_{t+1}}Q(s_{t+1},a_{t+1}) - Q(s_{t},a_{t}) ).
\end{equation}

In equation (2), $\alpha$ represents the learning rate, $r$ represents the reward for a given state and action, and $\gamma$ represents the discount factor. One can see that in the $\max_{a_{t+1}}Q(s_{t+1},a_{t+1})$ term, the idea is to determine the best possible future reward by taking this action. 

\subsection{Deep Q-Network (DQN)}

While Q-learning performs well in environments where the state-space is small, as the state-space begins to increase, Q-leaning becomes intractable. It is because there is now a need for more experience (more game episodes to be played) in the environment to allow convergence of the Q-values. To obtain Q-value estimates in environments where the state-space is large, the agent must now generalize from limited experience to states that may have not been visited \citep{Kochenderfer:2015:DMU:2815660}. One of the most widely used function approximation techniques for Q-learning is deep Q-networks (DQN), which involves using a neural network to approximate the Q-values for all the states. With standard Q-learning, the Q-value was a function of $Q(s,a)$, but with DQN the Q-value is now a function of $Q(s,a,\theta)$, where $\theta$ is the parameters of the neural network. Given an \textit{n}-dimensional state-space with an \textit{m}-dimensional action space, the neural network creates a map $R^{n} \rightarrow R^{m}$.  As mentioned by \citet{van2016deep}, incorporating a target network and experience replay are the two main ingredients for DQN \citep{mnih2015human}. The target network with parameters $\theta^{-}$, is equivalent to the on-line network, but the weights ($\theta^{-}$) are updated every $\tau$ time steps. The target used by DQN can then be written as:

\begin{equation}
Y_{t}^{DQN} = r_{t+1} + \gamma\max_{a_{t+1}}Q(s_{t+1},a_{t+1};\theta_{t}^{-}).
\end{equation}

The idea of experience replay is that for a certain amount of time, observed transitions are stored and then sampled uniformly to update the network. By incorporating the target network, as well as, experience replay, this can drastically improve the performance of the algorithm \citep{mnih2015human}.

\subsection{Double Deep Q-Network (DDQN)}

In Q-learning and DQN there is the use of a max operator to select which action results in the largest potential future reward. \citet{van2016deep} showed that due to this max operation, the network is more likely to overestimate the values, resulting in overoptimistic $Q$ value estimations. The idea introduced by \citet{hasselt2010double} was to decouple the max operation to prevent this overestimation to create what is called double deep Q-network (DDQN). To decouple the max operator, a second value function must be introduced, including a second network with weights $\theta'$. During each training iteration, one set of weights determines the greedy policy and the other then determine the Q-value associated. Formulating equation (2) as a DDQN problem:

\begin{equation}
Y_{t}^{DDQN} = r_{t+1} + \gamma Q(s_{t+1},\argmax_{a_{t+1}}Q(s_{t+1},a_{t+1};\theta_{t});\theta_{t}^{'}).
\end{equation}

In equation (3), it can be seen that the max operator has been removed and we are now including an $\argmax_{a_{t+1}}Q(s_{t+1},a_{t+1};\theta_{t})$ function to determine the best action due to the on-line weights. We then use that action, along with the second set of weights to determine the estimated Q-value. 

\section{Related Work}

Learning in complex, hierarchical environments is a challenging task and has attracted a lot of attention in recent publications \citep{sutton1999between,precup2000temporal, dayan1993feudal, dietterich2000hierarchical, boutilier1997prioritized, dayan1993improving, kaelbling1993hierarchical, parr1998reinforcement, precup1997planning, schmidhuber1991neural, sutton1995td, wiering1997hq, vezhnevets2016strategic, bacon2017option, vezhnevets2017feudal}. One popular method for constructing hierarchical agents is based off of work by \citep{sutton1999between, precup2000temporal}, known as the \textit{options} framework. In this framework, a low-level agent, known as the \textit{option}, can be thought of as a sub-policy generated by the top-level agent that operates in an environment until a given termination criteria is met. A top-level agent picks an option, given its own policy, thus creating a hierarchical structure. It was mentioned in work by \citet{vezhnevets2017feudal} that the options are typically learned using sub-goals and 'pseudo-rewards' that are explicitly defined \citep{sutton1999between, dietterich2000hierarchical, dayan1993feudal}. In recent publications, it was demonstrated that learning a selection rule among predefined options using deep neural networks (DNN) delivers promising results in challenging, complex environments like Minecraft and Atari \citep{tessler2017deep, kulkarni2016hierarchical, oh2017zero}; in addition, other research has shown that it is possible to learn options jointly with a policy-over-options end-to-end \citep{vezhnevets2017feudal, bacon2017option}. In the hierarchical formulation by \citet{vezhnevets2017feudal}, their framework used a top-level agent which produces a meaningful and explicit goal for the bottom level to achieve. In this framework, the authors were able to create sub-goals that emerge as directions in the latent space and are naturally diverse, which differs from the options framework.

A key difference between our approach and the aforementioned approaches is that in our proposal the main agent propagates information to the low-level \textit{nested} agent that becomes included in the state space of the nested agent. This decreases training time in complex environments and leads to significantly better performance (see section 5).

\section{Methods}

We now introduce the Deep Nested Agent framework, a variant of deep hierarchical reinforcement learning where the information from the top level agent is propagated to the nested agent in the form of state augmentation. In this section we discuss how the Deep Nested Agent algorithm is formulated and provide pseudo-code for the algorithm (see Algorithm 1).

\subsection{Nested Agents}

Nested Agents thrive in environments where there is explicit hierarchical structure. In these hierarchical environments, there are typically different sets of actions that can be decoupled into one main agent action set and one nested agent action set (or many nested agent action sets) that operate at different granularities of time or sequence. By doing this, the main agent can propagate information to the nested agent by adding an extra dimension to the state of the nested agent. For example, consider a two-level hierarchical environment with two actions sets: $A_{1}, A_{2}$. If we assume that the actions in $A_{1}$ operate at a slower time granularity, then we can assign $A_{1}$ to the main agent, and $A_{2}$ to the nested agent as follows:
\begin{center}
$a_{M} \in A_{1}$\\
$a_{N} \in A_{2}$,
\end{center}
where the subscript $M$ corresponds to the main agent and $N$ corresponds to the nested agent. Once the action set is defined, we can then construct the progression of information from the main agent to the nested agent. Consider state $s_{M}$ for the main agent. Given an action $a_{M}$, we can propagate the information to the nested agent as follows:
\begin{equation*}
s_{N} = [s_{M}, a_{M}].
\end{equation*}
This framework is also applicable with inputs of images, where the nested agent will consist of one extra dimension as compared to the main agent. By avoiding the creation of many agents for main agent to choose from, there is much less training required for the nested agent as compared to current deep hierarchical methods since we are only adding one dimension to the state of the nested agent. In many of the current hierarchical formulations, a top-level agent chooses an action $a \in [1,...n]$ which selects a specific agent, $agent_{a}$. By only adding an extra dimension, the nested agent is able to learn more efficiently in a given environment, as well as, achieve superior performance without having to add more agents, which leads to reduced memory consumption. Figure 3 shows the progress of information from the main agent to the nested agent and Algorithm 1 provides pseudo-code for the Deep Nested Agent framework. We can see the difference in time-granularity for the main agent and the nested agent in Algorithm 1 from the main agent choosing an action before entering into a iterative loop that only involves the nested agent. We assumed that the main agent only takes one decision before passing the information down to the nested agent, but in general, the main agent could take more than one action before passing the information down the the nested agent.

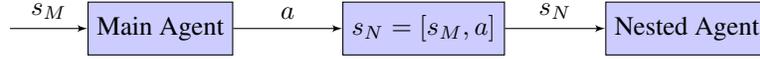
\begin{figure}[H]
\begin{center}
\tikzstyle{int}=[draw, fill=blue!20, minimum size=2em]
\tikzstyle{init} = [pin edge={to-,thin,black}]

\begin{tikzpicture}[node distance=2.5cm,auto,>=latex']
    \node [int] (a) {Main Agent};
    \node (b) [left of=a,node distance=2cm, coordinate] {a};
    \node [int] (c) [right of=a,node distance=3.5cm] {$s_{N} = [s_{M},a]$};
    
    \node [coordinate] (end) [right of=c, node distance=2cm]{};
	\node [int] (d) [right of=c,node distance=3.5cm] {Nested Agent};
    \path[->] (b) edge node {$s_{M}$} (a);
    \path[->] (a) edge node {$a$} (c);
    \path[->] (c) edge node {$s_{N}$} (d);
\end{tikzpicture}
\caption{Progression from main agent to nested agent.}
\end{center}
\end{figure}

\subsection{Exploration Vs. Exploitation}
In our experiments using the Deep Nested Agent framework, we implemented the $\epsilon$-$greedy$ search strategy. The problem with $\epsilon$-$greedy$ in a hierarchical reinforcement learning formulation is that if $\epsilon$ for the main agent decays at the same rate as $\epsilon$ for the nested agent, the convergence will be much slower since the main agent operates at a different time granularity. By allowing $\epsilon$ to decay faster for the main agent, this allows the main agent to take more \textit{greedy} actions sooner, ensuring that the nested agent has more time to explore with the state containing the \textit{greedy} main agent action. We also set the lower bound on $\epsilon$ to be greater for the main agent than that of the nested agent so that there is still a probability of exploring the other actions of the main agent and their effect on the nested agent.
\begin{algorithm}[H]
   \caption{Deep Nested Agent}
   \label{alg:NestedAgent}
\begin{algorithmic}
   \STATE {\bfseries Initialize:} Main Agent
   \STATE {\bfseries Initialize:} Nested Agent
   \STATE {\bfseries Initialize:} $s_{M}$
   \STATE reward = 0
   \STATE number of episodes = $n$
   \FOR{$i=1$ {\bfseries to} $n$}
   \STATE $a_{M} = ChooseAction$(Main Agent)
   \STATE $s_{N} = [s_{M}, a_{M}]$
   \REPEAT
   \STATE $a_{N} = ChooseAction$(Nested Agent)
   \STATE $s', r_{N} = SimulateEnvironment$
   \STATE receive main agent reward = $r_{M}(s')$
   \STATE $reward = reward + r$
   \STATE $update$(Nested Agent)
   \UNTIL{Terminal}
   \STATE $update$(Main Agent)
   \ENDFOR
\end{algorithmic}
\end{algorithm}
\section{Experiments}

We experimented with the Nested Agent framework on three non-Markovian scenarios within the Minecraft environment that consisted of constructing different designs of varying difficulty: a vertical line, a zigzag, and a  diamond (see Figure 4). In each scenario, the agent is only able to look in one direction: forward. This increases the complexity of the problem, because the agent can not look in a different direction and place a block, the agent can only place a block in front of where the agent is located. This is because we are currently using the Java version of Minecraft and are not using an API to interact with Minecraft. The blocks can be composed of two materials; wood or stone, and there is an associated penalty reward for selected either of the two. The non-Markovian element of the environment arises from the fact that the agent can only select to change material at the beginning of the episode and once the material is selected, it cannot be changed. Therefore, a decision early on in the environment (changing material) will have an effect later on that is not noticeable in the beginning. Our arena was composed of a 15 by 15 grid where each block in the grid corresponds to a block in Minecraft.

\begin{figure}[H]
\centering
\begin{subfigure}{.5\textwidth}
  \centering
  \includegraphics[width=.6\linewidth]{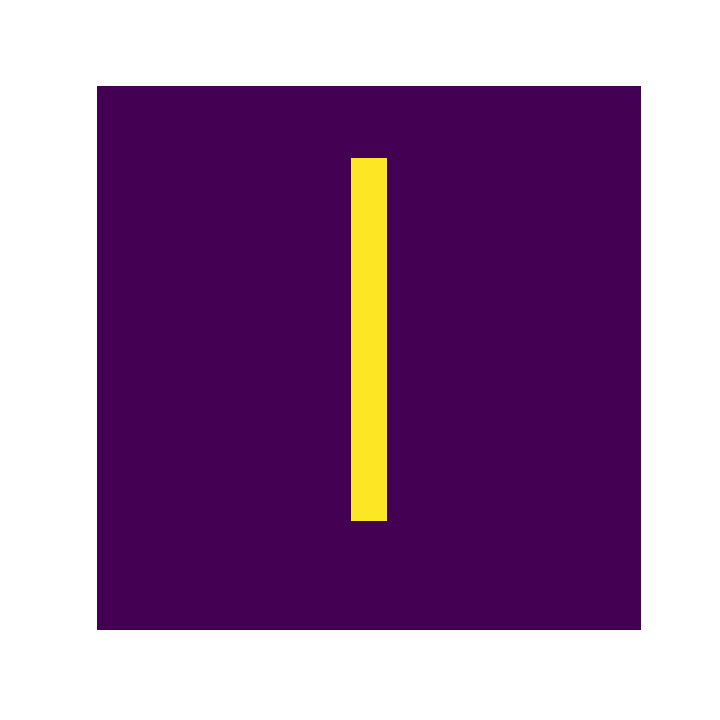}
  \caption{Scenario 1: Constructing a vertical line.}
  \label{fig:sub1}
\end{subfigure}%
\begin{subfigure}{.5\textwidth}
  \centering
  \includegraphics[width=.6\linewidth]{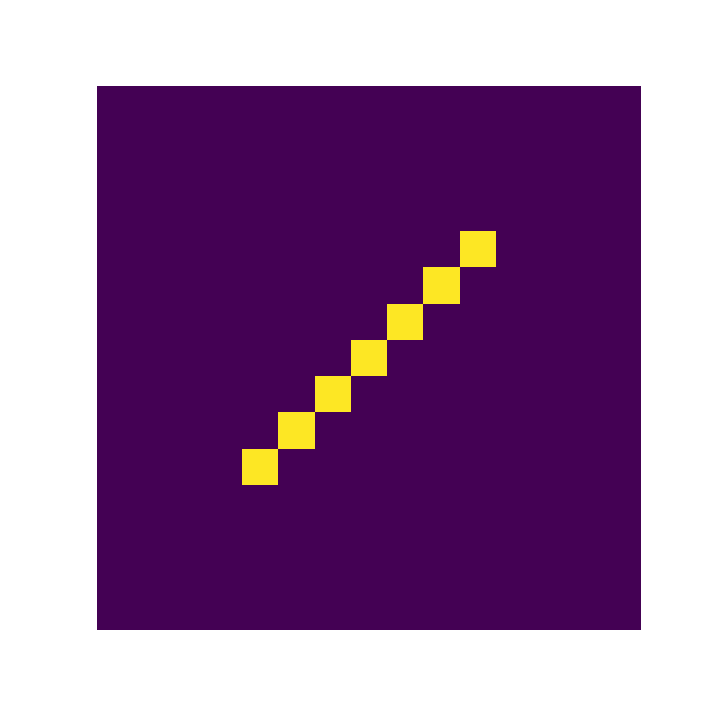}
  \caption{Scenario 2: Constructing a zigzag.}
  \label{fig:sub2}
\end{subfigure}
\begin{subfigure}{.5\textwidth}
  \centering
  \includegraphics[width=.6\linewidth]{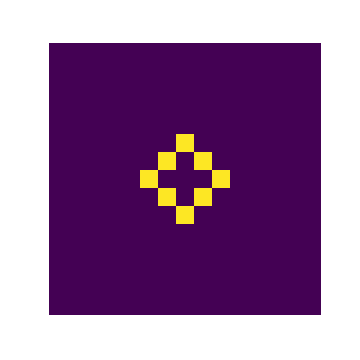}
  \caption{Scenario 3: Constructing a diamond.}
  \label{fig:sub2}
\end{subfigure}
\caption{Designs for the different scenarios.}
\label{fig:test}
\end{figure}

\subsection{Implementation}

\subsubsection{State-Space}

A state contains all the information the agent needs to make decisions. In our framework, the state-space of the main agent is different from the state of the nested agent. For the main agent, the information included in the state was the agent's position ($x_{i}$, $y_{i}$) and the number of available materials remaining (b). The reason for including the number of available materials is because when all materials were used, the episode terminated and a new one began. In each scenario there was an upper-bound on the number of available materials that was equal to the number of required materials to build the design. For the nested agent, the state space included the same information but added on the action of the main agent ($a_{M}$).

\subsubsection{Action-Space}
At each time-step, the main agent and the nested agent can make a decision to choose what material to use, as well as, move and drop blocks, respectively. The only difference is the the decision time-step for the main agent and nested agent. The main agent takes one action every episode, where an episode is defined as one entire evolution of the environment. The nested agent takes actions throughout the episode once the main agent has selected their action. The action-space for the main agent can be defined as follows:
\begin{equation*}
A_{M} = [Wood Blocks, Stone Blocks],
\end{equation*}

along with the action-space for the nested agent:

\begin{equation*}
A_{N} = [F,B,L,R,F+D,L+D,R+D,B+D],
\end{equation*}

where $F$ is move forward, $B$ backward, $L$ left, $R$ right, and the addition of $D$ means to place a block down.

\subsubsection{Reward Function}

The reward function for the main agent and nested agent needed to be designed to reflect the goal of the scenarios: constructing specific designs. We were able to capture our goals in the following reward functions for the main agent and the nested agent:

\paragraph{Main Agent}

\begin{equation*}
r_{M} = \sum_{i=1}^{15}\sum_{j=1}^{15} I(x_{i},y_{j}) + p,
\end{equation*}

\paragraph{Nested Agent}

\begin{equation*}
r_{N}(x_{i}, y_{j}) = I(x_{i},y_{j}),
\end{equation*}

where $I$ is an indicator function defined as:

\begin{equation*}
I(x_{i}, y_{j}) := \left\{\begin{array}{l}1;\quad if \; K(x_{i},y_{j}) = Shape(x_{i}, y_{j})\\0; \quad otherwise\end{array}\right.
\end{equation*}

and $p$ is a constant that depends on the action of the main agent. If the main agent chooses the wood material, $p = -5$ and if the main agent chooses the stone material, $p = 10$. We also defined two (15x15) matrices $K(x_{i},y_{j})$ and $Shape(x_{i}, y_{j})$, where $K$ contains the locations that the agent placed a block down and $Shape$ contains the locations of where the blocks should be placed from the scenario design (see Figure 4).

\subsection{Nested Agent Performance}

Our deep nested agent framework outperformed the deep hierarchical framework, as well as, the deep non-hierarchical single agent framework in terms of training stability, converged score, and computational expense. A detailed explanation of the deep non-hierarchical single agent framework along with the deep hierarchical framework are provided in the Appendix. The performance averaged over 10 trials through training is shown in Figure 5, where we evaluated the model after every 30 episodes. In all scenarios we allowed the agent to learn for 3,000 episodes. We can see how the deep nested agent framework was able to achieve much better performance early on on in scenario 1 and was ultimately able to achieve a greater converged score. In scenario 2 and scenario 3, the performance of the deep nested agent and hierarchical agent were comparable, but with the deep nested agent algorithm there was a benefit of not having to train an additional neural network. Both the deep nested agent framework and the deep hierarchical framework outperformed the deep non-hierarchical single agent framework in all scenarios. By using this nested agent framework, we avoided having to train separate neural networks for each action of the main agent, which was less computationally expensive.

\begin{figure}[H]
\centering
\begin{subfigure}{.5\textwidth}
  \centering
  \includegraphics[width=.8\linewidth]{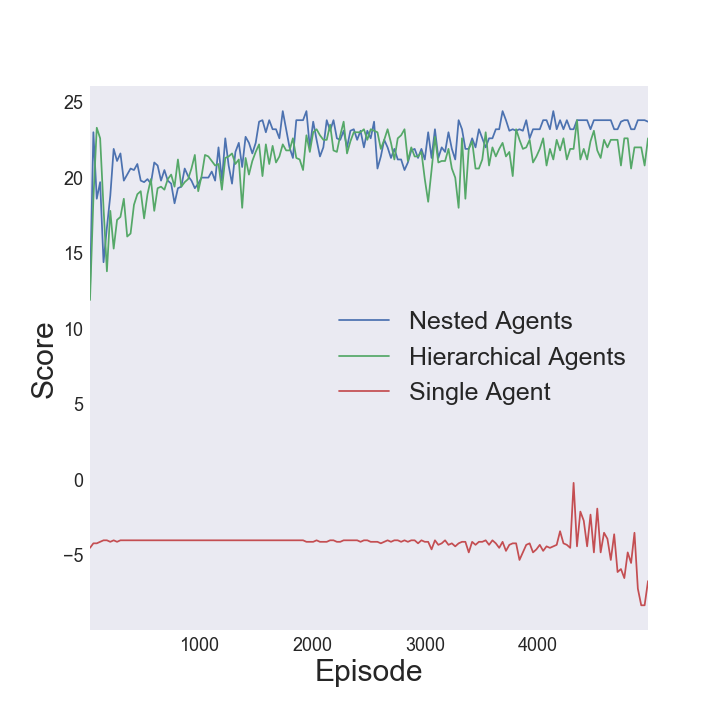}
  \caption{Scenario 1.}
  \label{fig:sub1}
\end{subfigure}%
\begin{subfigure}{.5\textwidth}
  \centering
  \includegraphics[width=.8\linewidth]{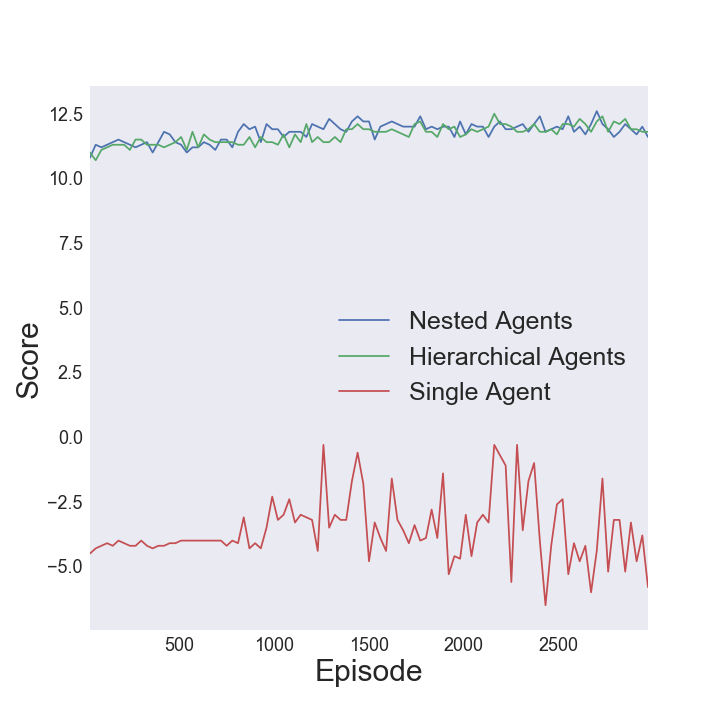}
  \caption{Scenario 2.}
  \label{fig:sub2}
\end{subfigure}
\begin{subfigure}{.5\textwidth}
  \centering
  \includegraphics[width=.8\linewidth]{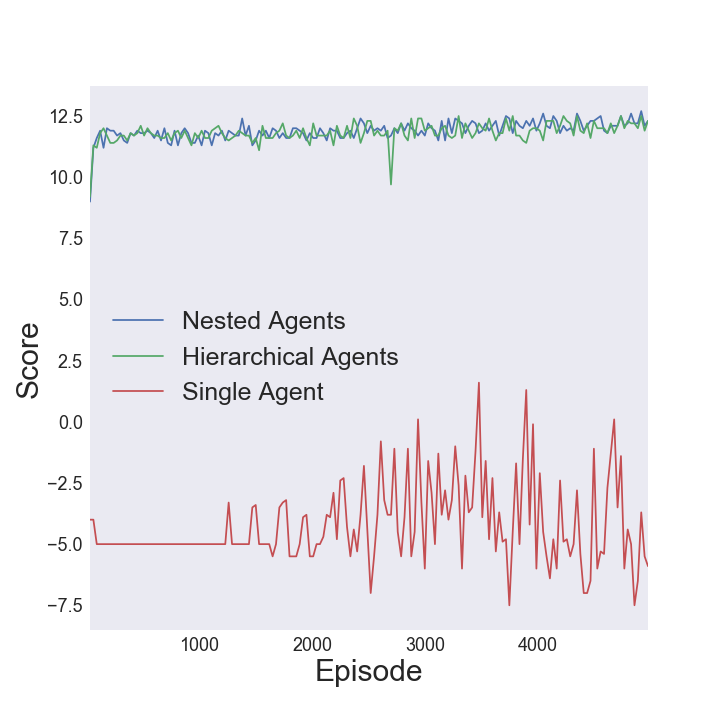}
  \caption{Scenario 3.}
  \label{fig:sub2}
\end{subfigure}
\caption{Score throughout training on the scenarios.}
\label{fig:test}
\end{figure}

\section{Conclusion and Future Work}
We introduced the deep nested agent framework, a variant of deep hierarchical reinforcement learning where the action of the main agent is included in the state of the nested agent. We found that the performance of the deep nested agent framework outperforms the deep hierarchical framework, as well as, the non-hierarchical single agent framework in environments that exhibit non-Markovian properties based on the stability of learning, the computational complexity, and the converged score in the scenarios. We also found that we reduced the computational demand for this problem by eliminating the need for training additional neural network models for each action of the main agent. By adding the action of the main agent to the state of the nested agent, we only had to train two DDQN models, compared with three for the deep hierarchical framework (this value increases for each unique main agent action). As future work, we are applying this framework to multi-agent scenarios where there are many nested agents that are cooperating with each other. We are particularly interested in this problem since the convergence in multi-agent scenarios is not guaranteed and we believe the deep nested agent framework will be able to improve the current state-of-the-art results. In addition, we plan to apply this framework to more complex environments where the main agent has many actions to take. In these environments, as the number of main agent actions increases, we believe the performance will increase as well, when comparing to the deep hierarchical framework and the deep non-hierarchical single agent framework. We envision our deep nested agent framework will soon be able to solve more complex problems, as well as, achieve better performance while reducing the required computational resources.

\bibliographystyle{unsrtnat}
\bibliography{nips.bib}
\newpage
\section{Appendix}

\subsection{Experiment Setup and Results}

All code for the experiments were run in python 3.6 using keras with tensorflow compiled from source. The experiments were run on a Ubuntu workstation with a single 12gb memory 1080x TI GPU with 32gb of RAM. We used the JAVA version of Minecraft, so all code had to be designed to know where to click on the game screen to start, with all actions being programmed as time-interval key strokes. As mentioned in the paper, we ran each model (non-hierarchical single agent, hierarchical agent, and nested agent) for 3,000 episodes and repeated the training for 10 trials to average the results. This led to a more interpretable result instead of reporting one single trial. I can be seen that the models were unable to obtain the optimal solution in scenario 2 and scenario 3. We believe this is due to the number of episodes being set to 3,000 and with more training the deep nested agent model and the deep hierarchical agent model should be able to obtain the optimal solution.

\subsubsection{Deep Non-Hierarchical Single Agent}
To construct the deep non-hierarchical single agent, we utilized the DDQN concept that was mentioned earlier in the paper. Since the agent did not have any hierarchical properties, the action set of the main agent and nested agent were combined into one action set for the single agent. Because the action of choosing material was time-sensitive, if the single agent did not choose the material first, then the episode would terminate.

\subsubsection{Deep Hierarchical Agent}
In the deep hierarchical agent algorithm, we had three total agents: one top-level agent and two low-level agents. The top-level agent's action was to select the low-level agent to construct the scenario's design. Each low-level agent was assigned a material to build, so one agent could build with stone and the other agent could build with wood. No information of the top-level agent was included in the state of the low-level agent and in this scenario we had to train an additional neural network since we have two low-level agents instead of one like in the deep nested agent algorithm.

\subsection{Hyper-parameters}
We used an experience replay memory length of one-million with no priority, but in the future we will compare our results using prioritized experience replay. Our neural network model was composed of a simple 4-layer network with two hidden layers. Each hidden layer had 32 nodes and we used a batch size of 32 as well. In all of the experiments we used the $TanH$ activation function in training our neural network. Each agent was trained using the DDQN architecture that was introduced in Section 2 with the addition of the deep nested agent framework. We also used the $\epsilon-greedy$ search strategy with $\epsilon$ decaying faster for the main agent and $\epsilon$ decaying slower for the nested agent. $\epsilon$ was linearly decayed from 1.0 $\rightarrow$ 0.001 in all experiments for the main agent and nested agent. 

\end{document}